\documentclass[10pt,twocolumn]{article}

\usepackage[margin=0.72in]{geometry}
\usepackage{times}
\usepackage{microtype}
\usepackage{graphicx}
\usepackage{amsmath,amssymb,mathtools}
\usepackage{booktabs}
\usepackage{multirow}
\usepackage{enumitem}
\usepackage{xcolor}
\usepackage{url}
\usepackage{natbib}
\usepackage[hidelinks]{hyperref}
\usepackage{caption}
\usepackage{tcolorbox}
\usepackage{placeins}

\definecolor{keygray}{gray}{0.96}
\definecolor{keyline}{gray}{0.55}

\newcommand{\rqfigure}[3]{%
\begin{figure}[t]
    \centering
    \IfFileExists{#1}{%
        \includegraphics[width=\linewidth]{#1}%
    }{%
        \fbox{\parbox[c][1.45in][c]{0.94\linewidth}{\centering
        \textbf{Figure placeholder}\\[2pt]
        Upload the missing figure}}%
    }
    \caption{#2}
    \label{#3}
\end{figure}
}

\newcommand{\rqfigurewide}[3]{%
\begin{figure*}[t]
    \centering
    \IfFileExists{#1}{%
        \includegraphics[width=0.92\textwidth]{#1}%
    }{%
        \fbox{\parbox[c][1.65in][c]{0.88\textwidth}{\centering
        \textbf{Figure placeholder}\\[2pt]
        Upload the missing figure}}%
    }
    \caption{#2}
    \label{#3}
\end{figure*}
}

\newtcolorbox{keybox}{
    colback=keygray,
    colframe=keyline,
    boxrule=0.45pt,
    arc=2pt,
    left=5pt,right=5pt,top=4pt,bottom=4pt
}

\newcommand{\key}[1]{\textbf{#1}}

\title{\vspace{-0.3in}
\textbf{Beyond Fixed Directions:}\\
Adaptive Representation Analysis of Reasoning and Memorization in LLMs
}

\author{
Shaheen Nabi\\
Independent Researcher\\
\texttt{devshaheen0@gmail.com}
}

\date{}

\begin{document}
\maketitle

\begin{abstract}
Recent work has proposed that reasoning and memorization in language models can be characterized by a single representation direction, including methods that keep this direction fixed during reinforcement learning. We test two assumptions behind this view. First, are reasoning-oriented and factual-recall task groups approximately single-direction separable? Second, does the resulting geometry remain stable after GRPO? Using Qwen3-0.6B and a controlled 400-example dataset, we find that a one-dimensional projection can match a full 1024-dimensional linear probe with AUROC \(=1.00\) on the studied task groups. However, after GRPO, the corresponding direction is substantially reorganized: mean-direction cosine averages \(0.453\), probe-direction cosine \(0.445\), while direct representation drift reaches \(0.511\) at the final layer. Probe AUROC nevertheless remains \(1.00\). The evidence therefore supports \key{single-direction decodability} for the studied task groups but challenges \key{fixed-direction stability}: the information persists while its geometric realization changes.
\end{abstract}

\begin{keybox}
\textbf{Central result.}
\emph{The reasoning--factual-recall distinction remains highly decodable after RL, but the specific direction carrying that distinction is not stable.}
This separates \textbf{information preservation} from \textbf{geometric preservation}.
\end{keybox}

\section{Introduction}

Reinforcement learning has become a central mechanism for eliciting stronger reasoning behavior in large language models, particularly through verifier-based optimization and group-relative objectives such as GRPO \citep{shao2024deepseekmath,guo2025deepseekr1}. In parallel, mechanistic representation studies have asked whether high-level behavioral distinctions correspond to simple directions in activation space. Most directly related to this work, Hong et al.~\citep{hong2025direction} report that reasoning- and memorization-intensive inputs can be separated along a single direction, and DiRL uses such a direction as a fixed geometric anchor during RL \citep{xia2026dirl}.

The practical appeal of a fixed direction is clear: a representation-level signal can be extracted once and reused during optimization. But this design implicitly raises a stability question. A direction may be highly predictive before RL without remaining aligned with the representation after the model's parameters have changed. Thus, \textbf{linear separability} and \textbf{directional stability} are distinct empirical properties.

We study these properties in Qwen3-0.6B through two focused research questions:

\begin{itemize}[leftmargin=1.1em,itemsep=1pt,topsep=2pt]
    \item \key{RQ1 -- Separability:} Can the reasoning-oriented versus factual-recall distinction be captured by a single linear direction across transformer layers?
    \item \key{RQ2 -- Stability:} Does that representation remain geometrically aligned after GRPO, or does reinforcement learning reorganize the representation?
\end{itemize}

The study deliberately stops at these two questions. We do not attempt to establish cross-task transfer, sparse-autoencoder decomposition, or causal mechanistic replacement. The resulting contribution is narrower: \textbf{a controlled representation-level test of the fixed-direction assumption}.

\paragraph{Contributions.}
We make three focused contributions. First, we strengthen the RQ1 control protocol by moving from the original dataset comparison to a 400-example, length-matched multi-dataset evaluation. Second, we directly compare mean-difference and learned probe directions across all 28 transformer layers. Third, we compare the base representation with a public GRPO-trained checkpoint and show that near-perfect decodability coexists with substantial directional rotation and representation drift.

\section{Prior Work and the Fixed-Direction Assumption}

\subsection{The Linear Representation Hypothesis}

The linear representation hypothesis motivates the study of semantic and behavioral properties through directions in activation space \citep{park2023linear}. Empirical work has demonstrated that transformer representations often expose approximately linear relational structure \citep{hernandez2024linearity}. Under this hypothesis, a high-level behavioral or semantic property is expected to correspond, at least approximately, to a single direction \(\hat d\in\mathbb{R}^{D}\) such that the scalar projection \(h(x)^{\top}\hat d\) is monotonically related to the presence of the property in \(x\). Park et al.~\citep{park2023linear} formalize this as a statement about residual-stream geometry rather than a claim about any specific layer, which is precisely the property our RQ1 protocol tests empirically at the layer level. Hernandez et al.~\citep{hernandez2024linearity} independently show that relation decoding is often well approximated by a single affine transformation, reinforcing the plausibility of a linear-direction account here. Our contribution is a controlled empirical test of whether one specific behavioral distinction -- reasoning-oriented versus factual-recall task framing -- satisfies this hypothesis, and whether that satisfaction persists across a training intervention.

\subsection{Reasoning versus Memorization in Language Models}

A parallel line of work asks whether language models solve problems through generalizable computation or retrieval of memorized patterns. Dziri et al.~\citep{dziri2023faith} show that high performance on compositional tasks can mask reliance on pattern matching that fails under systematic perturbation, and Berglund et al.~\citep{berglund2024reversal} document a related asymmetry in which models that learn ``A is B'' frequently fail to infer ``B is A,'' suggesting facts are stored in a directionally specific, retrieval-like format. These behavioral results motivate the reasoning/factual-recall distinction we study, but do not themselves establish anything about representation geometry.

Hong et al.~\citep{hong2025direction} provide the most direct precedent for the present analysis by arguing that reasoning--memorization interplay is mediated by a single direction. DiRL subsequently operationalizes this representation as a fixed anchor for direction-aware exploration during GRPO \citep{xia2026dirl}, explicitly computing the initial direction from the residual stream and keeping it fixed during RL. Our RQ1 protocol is consistent with, and complementary to, the empirical claim in Hong et al.~\citep{hong2025direction}: we do not dispute that a single direction can separate reasoning-oriented from factual-recall inputs. Where our study diverges is in asking a question a single-model-state analysis cannot answer by construction -- whether the direction identified before training remains the same direction after training.

\subsection{Mechanistic Analyses and Sparse Decompositions}

Hou et al.~\citep{hou2023mechanistic} investigate internal signatures of multi-step reasoning using attribution and ablation, providing evidence that reasoning-relevant computation is at least partially localizable. Sparse autoencoder (SAE) decompositions have further shown that individual activation directions can correspond to interpretable, near-monosemantic concepts \citep{bricken2023towards}, and representation engineering uses extracted directions as steering vectors to control behavior at inference time \citep{zou2023representation}. Our object of study differs from this program: rather than asking whether a direction can be found or used for control, we ask whether a direction found before RL remains geometrically valid after RL.

\subsection{Probing and Control Tasks}

Our use of a full 1024-dimensional logistic-regression probe as a reference classifier follows standard probing methodology \citep{hewitt2019designing}. Hewitt and Liang~\citep{hewitt2019designing} caution that probe accuracy alone can overstate the information genuinely present in a representation, since an expressive probe can partially memorize labels through control-task-like shortcuts. We address this with our random-label probe control (Section~\ref{sec:controls}), and the near-identical performance of the one-dimensional constructions and the full probe (Table~\ref{tab:rq1-main}) is evidence against a probe-capacity artifact, since the full probe has far more capacity to memorize idiosyncratic patterns than a single direction yet does not diverge from it.

\subsection{Representation Similarity Across Model States}

A separate literature compares neural network representations directly, most notably through centered kernel alignment (CKA) \citep{kornblith2019similarity}, typically applied across architectures or layers as a population-level statistic. Our representation-drift measurement (Section~\ref{sec:repr-drift}) is conceptually related but differs in computing per-example cosine similarity of matched representations across two checkpoints of the same architecture. We view per-layer drift and CKA as complementary tools; a full CKA-based comparison is left to future work.

\subsection{Reinforcement Learning for Reasoning: GRPO}

Group-relative policy optimization (GRPO) has become a central post-training mechanism for eliciting stronger reasoning behavior in open language models, most prominently in DeepSeekMath \citep{shao2024deepseekmath} and DeepSeek-R1 \citep{guo2025deepseekr1}. GRPO replaces a learned value function with a group-normalized advantage estimate, substantially reducing training cost relative to PPO-style RLHF while still producing large reasoning gains. Because GRPO updates all parameters through gradient descent on a policy objective, it provides no a priori guarantee that a representational direction identified before training remains fixed -- the empirical gap RQ2 closes for the direction studied here.

\subsection{Fixed-Direction Interventions: DiRL}

DiRL~\citep{xia2026dirl} uses a reasoning-versus-memorization direction, extracted once from the base model, as a fixed geometric anchor to bias exploration during GRPO training. This is a practical, well-motivated design choice: a fixed anchor is cheap to compute and can be validated by its effect on downstream reward. Our results do not evaluate DiRL's downstream performance and should not be read as a critique of its practical utility (Section~\ref{sec:dirl-discussion}). Our contribution is to measure a property -- post-RL geometric stability of the anchor direction -- that is assumed but not directly tested by a fixed-anchor design.

\subsection{Positioning of This Work}

Our paper asks a narrower question about this design choice. We distinguish:

\begin{enumerate}[leftmargin=1.35em,itemsep=1pt,topsep=2pt]
    \item \key{Decodability}: can the task-group distinction be recovered from hidden states?
    \item \key{Direction stability}: does the same separating vector remain aligned after training?
    \item \key{Mechanistic stability}: does the same causal computation implement the distinction?
\end{enumerate}

We test the first two. We do \emph{not} claim to establish the third.

\section{Experimental Design}

\subsection{Problem Formulation}
\label{sec:formulation}

We first state the quantities used throughout the paper formally, since RQ1 and RQ2 are both defined directly in terms of them.

\paragraph{Hidden representation.} For an input example \(x\) and transformer layer \(\ell\in\{1,\dots,28\}\), let \(h_\ell(x)\in\mathbb{R}^{1024}\) denote the hidden state extracted at the final formatted prompt-token position of layer \(\ell\). A model checkpoint \(M\in\{B,\mathrm{RL}\}\) (base or GRPO-trained) induces a family of representation functions \(h_\ell^{M}(\cdot)\), so that the same input \(x\) yields two generally distinct vectors \(h_\ell^{B}(x)\) and \(h_\ell^{RL}(x)\) at a fixed layer \(\ell\).

\paragraph{Mean-difference direction.} Given a labeled dataset partitioned into reasoning-oriented examples \(D_R\) and factual-recall examples \(D_F\), the class mean-difference direction at layer \(\ell\) for model \(M\) is
\[
d_\ell^{M} = \frac{1}{|D_R|}\sum_{x\in D_R} h_\ell^{M}(x) \;-\; \frac{1}{|D_F|}\sum_{x\in D_F} h_\ell^{M}(x).
\]
This is the simplest possible estimator of a separating direction: it requires no optimization and is fully determined by first-order class statistics.

\paragraph{One-dimensional projection.} Given a candidate unit direction \(\hat d_\ell = d_\ell/\lVert d_\ell\rVert\), an example \(x\) is scored by the signed scalar projection
\[
s_\ell(x) = \big(h_\ell(x)-\mu_{\mathrm{train}}\big)^{\top}\hat d_\ell,
\]
where \(\mu_{\mathrm{train}}\) is the training-fold mean representation. Classification from \(s_\ell(x)\) alone -- via a single learned threshold -- is what we mean by a \emph{one-dimensional} or \emph{single-direction} classifier, as distinct from a full-rank linear probe.

\paragraph{Linear-probe direction.} We separately fit an unregularized-capacity logistic-regression probe \(w_\ell\in\mathbb{R}^{1024}\), \(b_\ell\in\mathbb{R}\) at each layer by minimizing binary cross-entropy over the training fold. The normalized probe weight vector \(\hat w_\ell = w_\ell/\lVert w_\ell\rVert\) provides a second, independently constructed one-dimensional direction: unlike \(\hat d_\ell\), it is obtained by discriminative optimization rather than by a class-conditional first moment, so agreement between \(\hat d_\ell\) and \(\hat w_\ell\) is not guaranteed by construction.

\paragraph{Direction stability.} For a direction estimator \(\hat v_\ell\in\{\hat d_\ell,\hat w_\ell\}\) computed independently on the base and RL checkpoints, we define \emph{direction stability} at layer \(\ell\) as the cosine similarity
\[
\operatorname{cos}\!\big(\hat v_\ell^{B},\hat v_\ell^{RL}\big) = \hat v_\ell^{B}\cdot \hat v_\ell^{RL} \in[-1,1].
\]
A value near \(1\) indicates that the same direction, in the same coordinate system, continues to separate the two task groups after training; a value near \(0\) indicates that the post-training separating direction is close to orthogonal to the pre-training one.

\paragraph{Representation drift.} Independently of any classifier or direction, we define the \emph{representation drift} at layer \(\ell\) as the example-averaged cosine distance between matched hidden states across checkpoints,
\[
\operatorname{drift}_\ell = 1 - \mathbb{E}_{x}\Big[\operatorname{cos}\big(h_\ell^{B}(x), h_\ell^{RL}(x)\big)\Big].
\]
This quantity makes no reference to the task labels at all: it measures how much the representation of the \emph{same input} has moved in activation space, independent of whether that movement happens to be aligned with the reasoning/factual-recall axis. Direction stability and representation drift are therefore logically independent quantities that could in principle disagree; Section~\ref{sec:repr-drift} and Section~\ref{sec:info-vs-geometry} report that, empirically, they agree closely.

With these four objects -- \(d_\ell\), \(\hat w_\ell\), direction stability, and representation drift -- RQ1 reduces to a comparison between one-dimensional classifiers using \(d_\ell\) or \(\hat w_\ell\) and the full probe \(w_\ell\) at a single checkpoint, and RQ2 reduces to tracking direction stability and representation drift across the base-to-RL transition.

\subsection{Models}

For RQ1, we use the custom Qwen3-0.6B base implementation used throughout the representation-analysis pipeline. For RQ2, we compare this base model with the publicly released GRPO-trained checkpoint \texttt{x32/Qwen3-0.6B-GRPO-GSM8K-Think}. The RL checkpoint was not trained in this study; it is treated as an independently available post-training state.

\subsection{Controlled Dataset}

The final RQ1/RQ2 analysis uses \textbf{400 examples}: 200 reasoning-oriented and 200 factual-recall examples. The reasoning group is drawn from MATH-500 and GSM8K; the factual-recall group is drawn from PopQA and TriviaQA. Exact duplicates and duplicate identifiers are removed before matching.

The original MATH-500 versus PopQA diagnostic experiment was intentionally treated as a confounded baseline. Token length alone achieved approximately \(0.965\) accuracy and \(0.994\) AUROC in that setting. A strict pairwise length-matched experiment reduced the confound but produced only 20 examples. We therefore use the expanded multi-dataset construction as the principal RQ1 dataset.

The raw source collection contains 160,624 examples: 500 MATH-500, 7,473 GSM8K, 14,267 PopQA, and 138,384 TriviaQA examples. After deduplication, 97,477 examples remain. Distributional matching using token-length bins of width 10 produces the final balanced 400-example set.

We consistently use the term \emph{factual recall} rather than equating factual recall with memorization. A factual question may be solved through retrieval, general knowledge, or reasoning; the label is therefore an operational task category rather than a causal claim about how the model acquired the information.

\subsubsection{From an Underpowered Pairwise Diagnostic to an Expanded Construction}

The dataset construction proceeded through three stages, each addressing a specific weakness in the previous one.

\emph{Stage 1 -- original diagnostic.} The initial experiment compared MATH-500 \citep{hendrycks2021math} directly against PopQA \citep{mallen2023popqa} without any length control, which is scientifically weak on its own: MATH-500 problems are systematically longer than PopQA questions, so a hidden-state classifier could be recovering prompt length rather than semantic distinction. We report this stage only as a motivating diagnostic (Appendix~\ref{app:original-diagnostic}), not as evidence for RQ1.

\emph{Stage 2 -- strict pairwise length matching.} We next attempted a strict pairwise match, pairing each reasoning example with a factual-recall example of nearly identical token length. This removes the confound but is highly data-inefficient, yielding only 20 examples -- too small for reliable held-out evaluation of a 1024-dimensional probe (Appendix~\ref{app:original-diagnostic}).

\emph{Stage 3 -- expanded, multi-dataset, distribution-matched construction.} We therefore broadened the source pool to four datasets -- MATH-500 \citep{hendrycks2021math}, GSM8K \citep{cobbe2021verifiers}, PopQA \citep{mallen2023popqa}, and TriviaQA \citep{joshi2017triviaqa} -- deduplicated the pool, and matched the two task-group distributions using token-length bins of width 10. Binned matching preserves far more data than strict pairwise matching while still equalizing length distributions at the population level (Figure~\ref{fig:rq1-length-matching}; Appendix~\ref{app:length-diagnostics}), and is the construction used for all main-text results.

\subsubsection{Dataset Composition}

Table~\ref{tab:dataset-composition} summarizes the raw and final composition of the four source datasets. GSM8K and MATH-500 jointly constitute the reasoning-oriented group; PopQA and TriviaQA jointly constitute the factual-recall group. TriviaQA and PopQA dominate the raw pool by a wide margin, which is precisely why naive sampling from the union of all four datasets would not, on its own, produce a length- or size-balanced comparison -- the explicit binning procedure in Stage 3 is what yields the final 400-example balanced set rather than simple down-sampling.

\begin{table}[t]
\centering
\small
\begin{tabular}{lccc}
\toprule
Dataset & Task type & Raw size & Final use\\
\midrule
GSM8K & Reasoning & 7{,}473 & length-binned\\
MATH-500 & Reasoning & 500 & length-binned\\
PopQA & Factual recall & 14{,}267 & length-binned\\
TriviaQA & Factual recall & 138{,}384 & length-binned\\
\midrule
\textbf{Total (raw)} & -- & \textbf{160{,}624} & --\\
\textbf{Total (deduplicated)} & -- & \textbf{97{,}477} & --\\
\textbf{Final balanced set} & -- & -- & \textbf{400}\\
\bottomrule
\end{tabular}
\caption{Composition of the source datasets used to construct the controlled RQ1/RQ2 evaluation set. The final 400-example set (200 reasoning-oriented, 200 factual-recall) is obtained by token-length-bin matching over the deduplicated pool, not by uniform random sampling.}
\label{tab:dataset-composition}
\end{table}

\subsection{Prompt and Representation Extraction}

All examples use the same formatting pipeline. Token lengths are measured after formatting with the Qwen3 tokenizer. For each example we extract the hidden representation at the final formatted prompt-token position from every transformer layer.

The resulting representation tensor has shape
\[
[400,\,28,\,1024],
\]
corresponding to 400 examples, 28 transformer layers, and a 1024-dimensional hidden state.

\subsection{RQ1: Direction Construction and Probing}

For layer \(\ell\), let \(h_\ell(x)\in\mathbb{R}^{1024}\) denote the extracted representation. We define the class mean-difference direction
\[
d_\ell =
\frac{1}{|D_R|}\sum_{x\in D_R} h_\ell(x)
-
\frac{1}{|D_F|}\sum_{x\in D_F} h_\ell(x),
\]
where \(D_R\) and \(D_F\) denote reasoning-oriented and factual-recall examples.

The scalar projection of an example is
\[
s_i=(h_i-\mu_{\mathrm{train}})^\top
\hat d_\ell,
\qquad
\hat d_\ell=\frac{d_\ell}{\|d_\ell\|}.
\]

We evaluate this one-dimensional signal with held-out stratified cross-validation. In parallel, we train a full 1024-dimensional logistic-regression probe independently at each layer. We also use the normalized probe weight vector as a second one-dimensional direction.

The key comparison is therefore:
\[
\text{one direction}
\quad\text{vs.}\quad
\text{full 1024-D probe}.
\]

If a one-dimensional direction matches the full probe, the observed task-group separation is empirically well described by a single direction, at least for the studied dataset and model.

\subsection{Controls}
\label{sec:controls}

We evaluate three important controls. First, a token-length classifier quantifies residual length information. Second, a TF--IDF lexical classifier quantifies how strongly surface vocabulary predicts the task label. Third, a random-label probe checks that the representation pipeline is not producing artificial separation.

These controls are important because high hidden-state AUROC alone does not establish semantic reasoning--memorization structure.

\subsection{RQ2: Direction Stability}

For each layer, we compare the base and RL mean-difference directions using
\[
\operatorname{cos}(d_\ell^{B},d_\ell^{RL})
=
\frac{d_\ell^{B}\cdot d_\ell^{RL}}
{\|d_\ell^{B}\|\,\|d_\ell^{RL}\|}.
\]

We independently train a linear probe for each model and compare their normalized probe weight vectors using the same cosine measure. This provides a second estimate of whether the separating geometry survives RL.

\subsection{Representation Drift}
\label{sec:repr-drift}

To quantify broader representational change, we compare the corresponding example representations:
\[
\operatorname{drift}_\ell
=
1-\mathbb{E}_{x}
\left[
\operatorname{cos}
\left(
h_\ell^{B}(x),h_\ell^{RL}(x)
\right)
\right].
\]

This measures geometric change. It is not, by itself, evidence that the underlying reasoning computation has causally changed.

\section{Results}

\subsection{RQ1: The Single-Direction Hypothesis}

The final RQ1 experiment provides strong support for an approximately single-direction separation of the \emph{studied task groups}. The best mean-difference direction occurs at layer 18 and reaches
\[
\text{Accuracy}=0.9975\pm0.0050,
\qquad
\text{AUROC}=1.0000\pm0.0000.
\]
The best probe-weight direction occurs at layer 12 and reaches the same AUROC, while the full 1024-dimensional probe also reaches
\[
\text{Accuracy}=0.9975\pm0.0050,
\qquad
\text{AUROC}=1.0000\pm0.0000.
\]

Thus, adding the remaining 1023 dimensions provides essentially no improvement in classification performance on this controlled dataset.

\begin{table}[t]
\centering
\small
\begin{tabular}{lccc}
\toprule
Method & Layer & Accuracy & AUROC\\
\midrule
Mean-difference direction & 18 & .9975 & 1.0000\\
Probe-weight direction & 12 & .9975 & 1.0000\\
Full 1024-D probe & 12 & .9975 & 1.0000\\
Random-label probe & -- & .4875 & .4816\\
\bottomrule
\end{tabular}
\caption{Final RQ1 results. A one-dimensional direction matches the full linear probe on the controlled task groups. Random labels produce chance-level performance.}
\label{tab:rq1-main}
\end{table}

\subsection{Length and Lexical Controls}

The control experiments substantially change how the original result should be interpreted. In the initial MATH-500 versus PopQA comparison, prompt length alone was highly predictive. The expanded 400-example construction reduces this dominant scalar confound, with length-only performance falling to
\[
0.6475\pm0.0533\ \text{accuracy},
\qquad
0.6645\pm0.0585\ \text{AUROC}.
\]

However, TF--IDF remains strong:
\[
0.9600\pm0.0146\ \text{accuracy},
\qquad
0.9911\pm0.0113\ \text{AUROC}.
\]

This is a central limitation rather than a result to hide. The hidden-state result demonstrates strong representational separation, but the task labels remain correlated with lexical and domain structure. Consequently, our claim is deliberately phrased as \textbf{approximately single-direction separation of the studied task groups}, not proof of a universal semantic reasoning axis.

\begin{table}[t]
\centering
\small
\begin{tabular}{lcc}
\toprule
Baseline / classifier & Accuracy & AUROC\\
\midrule
Length-only classifier & .6475 & .6645\\
TF--IDF lexical classifier & .9600 & .9911\\
Random-label probe & .4875 & .4816\\
Mean-difference direction (best layer) & .9975 & 1.0000\\
Probe-weight direction (best layer) & .9975 & 1.0000\\
Full 1024-D probe (best layer) & .9975 & 1.0000\\
\bottomrule
\end{tabular}
\caption{Complete baseline comparison for the RQ1 controlled task-group distinction. The length-only baseline is substantially weakened relative to the original unmatched diagnostic (Appendix~\ref{app:original-diagnostic}), while TF--IDF remains a strong but strictly weaker predictor than any hidden-state-based classifier. The random-label probe establishes the chance floor.}
\label{tab:baselines}
\end{table}

Table~\ref{tab:baselines} places all controls and all classifiers on a common scale. Three comparisons are informative. First, length-only accuracy falls from \(\approx\!0.965\) in the original unmatched diagnostic (Appendix~\ref{app:original-diagnostic}) to \(0.6475\) after binned length matching, confirming that the matching procedure meaningfully -- though not completely -- reduces the length confound. Second, TF--IDF remains considerably stronger than length alone (\(0.9600\) vs.\ \(0.6475\) accuracy), since lexical vocabulary differences between mathematical and factual-recall prompts persist even when token counts are equalized. Third, every hidden-state-derived classifier -- the two one-dimensional directions and the full probe -- exceeds the TF--IDF baseline and reaches near-ceiling performance, indicating the hidden-state representation carries strictly more separating information than surface lexical statistics alone, even though it does not eliminate the lexical correlation as an explanatory confound.

\subsubsection{Why Matching the Full Probe Is Scientifically Meaningful}

The central empirical claim of RQ1 is not merely that a one-dimensional signal is predictive, but that it is \emph{exactly as predictive as the full-rank alternative}. A 1024-dimensional probe has, in principle, the capacity to exploit any linear combination of feature directions, including combinations a simple mean-difference or single discriminative direction would miss. If the reasoning/factual-recall distinction were encoded across several weakly correlated directions -- length, lexical/topic content, and abstract task format, for instance -- we would expect the full probe to outperform any single projection by combining them. The fact that the one-dimensional constructions match the full probe's accuracy and AUROC to four decimal places at their respective best layers (Table~\ref{tab:rq1-main}) is evidence against such a multi-directional encoding, rather than merely evidence that \emph{some} linear signal exists.

This claim's scope is deliberately narrow: matching the full probe establishes \emph{effective} dimensionality one for classification on the studied 400-example set; it does not establish that the relevant subspace is exactly one-dimensional for all task-group definitions, datasets, or models.

\rqfigurewide{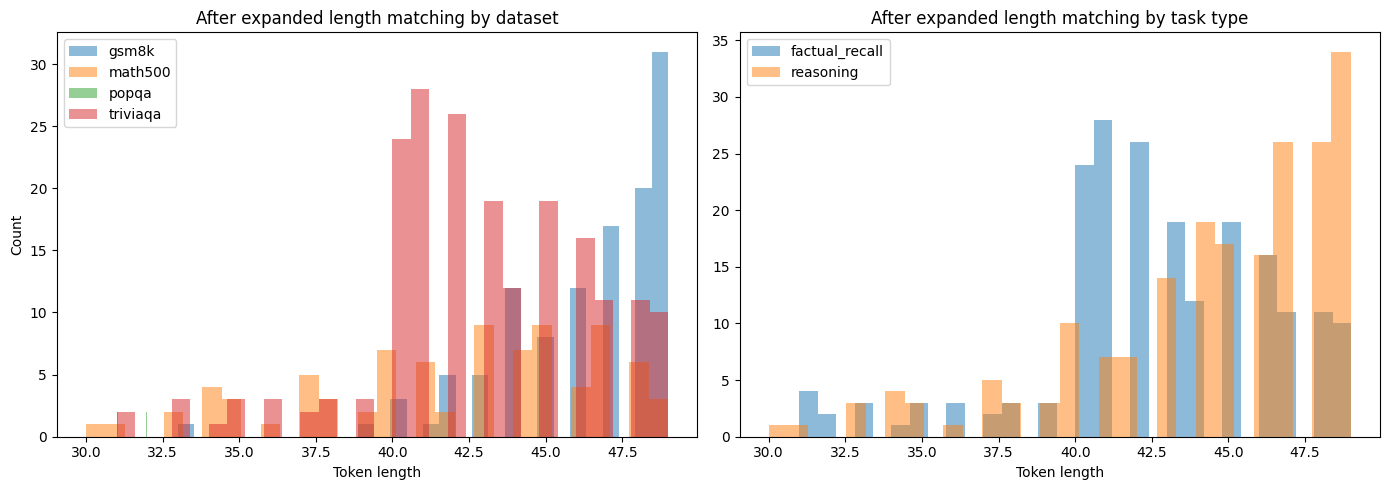}
{Expanded length-controlled token-length distributions after matching, shown by source dataset (left) and by task type (right). The final balanced dataset is substantially more controlled for prompt length than the original dataset comparison.}
{fig:rq1-length-matching}

Beyond the length control itself, a low-dimensional visualization of the resulting hidden-state geometry is informative before turning to the quantitative single-direction analysis. Figure~\ref{fig:rq1-pca} projects the final-layer representations of the controlled 400-example set onto their top two principal components: reasoning-oriented and factual-recall examples occupy visually distinct regions of this two-dimensional projection, which is consistent with -- though not sufficient on its own to establish -- the approximately single-direction separability quantified in Table~\ref{tab:rq1-main} and Section~\ref{sec:formulation}.

\rqfigurewide{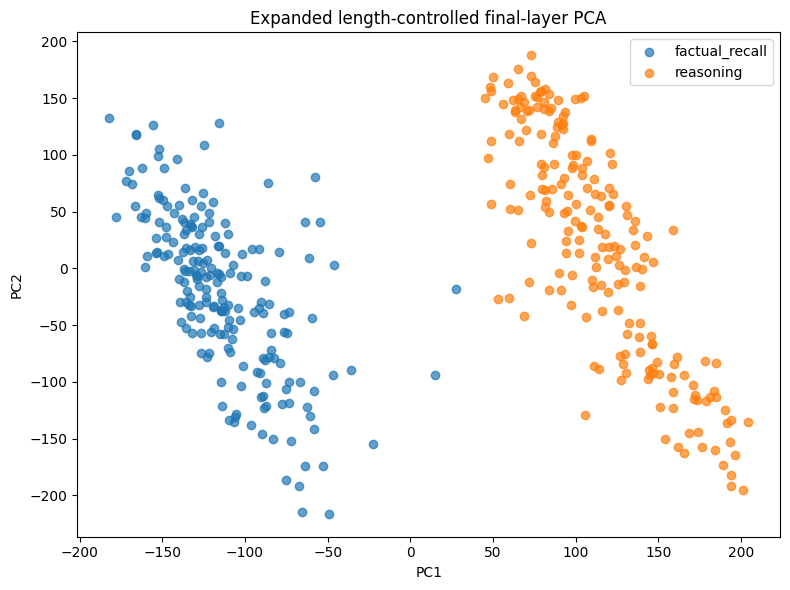}
{Expanded length-controlled final-layer PCA. Reasoning-oriented and factual-recall representations form two visually separated clusters along the leading principal components of the final-layer hidden state, consistent with an approximately linear separating structure on the controlled 400-example dataset.}
{fig:rq1-pca}

\subsection{Layer-wise Single-Direction Performance}

Both one-dimensional directions remain highly predictive across the 28 transformer layers. The mean-difference direction begins around AUROC \(0.992\) and reaches perfect AUROC from approximately layer 18 onward. The probe-weight direction begins around AUROC \(0.994\) and reaches perfect AUROC by approximately layer 12.

\rqfigure{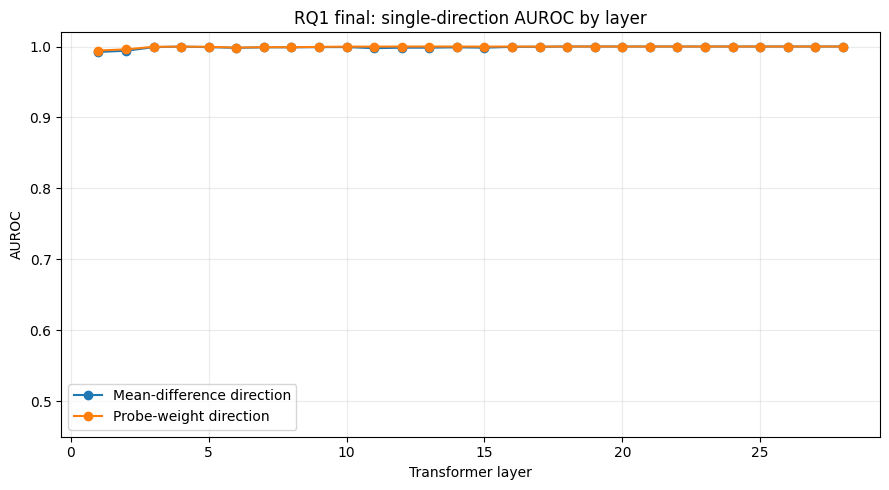}
{Layer-wise AUROC for the mean-difference and probe-weight directions. Both one-dimensional directions remain near-perfect across depth.}
{fig:rq1-single-direction}

The full 1024-dimensional probe is also essentially perfect across layers. Its role is primarily a reference: the important observation is that the explicit one-dimensional constructions match this richer classifier.

\rqfigure{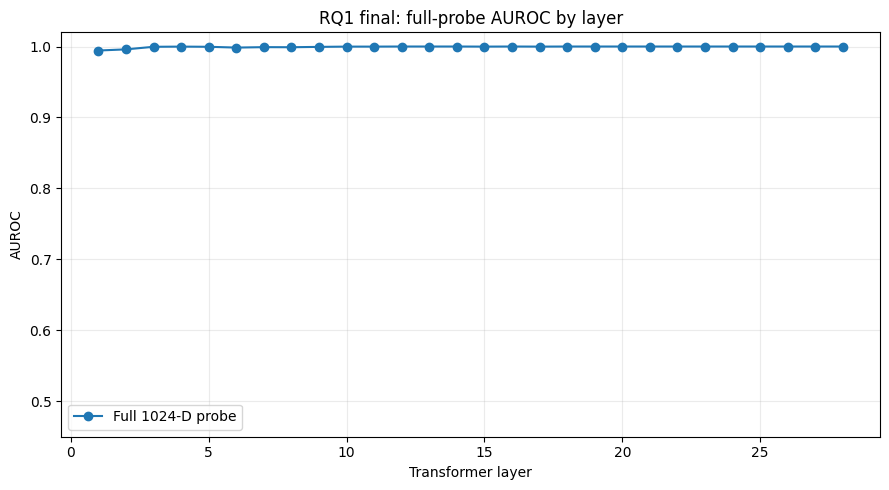}
{Layer-wise AUROC of the full 1024-dimensional linear probe. The full probe provides the high-dimensional reference against which one-dimensional directions are evaluated.}
{fig:rq1-full-probe}

\subsection{Direction Agreement and Score Geometry}

The mean-difference and probe-weight directions are strongly aligned. Their average cosine similarity across layers is approximately \(0.8495\). Alignment is highest in early layers, decreases through the middle of the network, and rises again in later layers.

\rqfigure{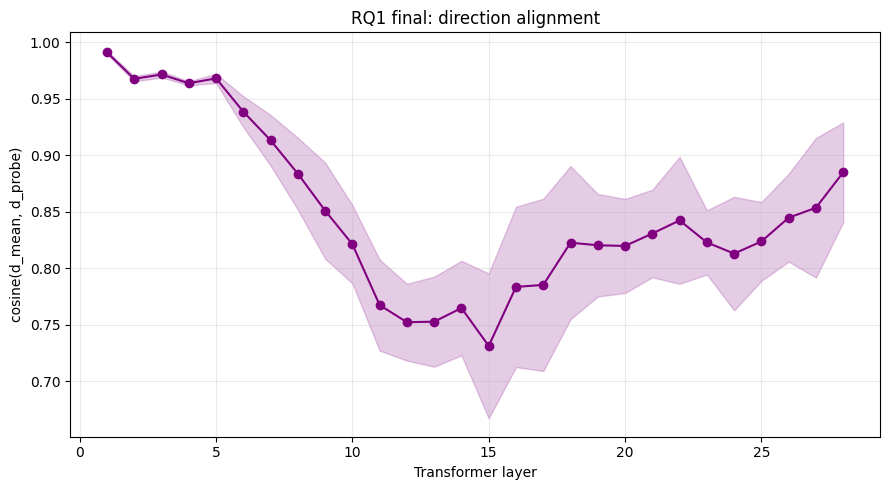}
{Cosine alignment between the class mean-difference direction and the independently learned probe-weight direction across transformer layers.}
{fig:rq1-alignment}

This agreement is scientifically important because \(d_\ell\) and \(\hat w_\ell\) are constructed by entirely different procedures. The mean-difference direction is a closed-form first-moment statistic with no optimization; the probe-weight direction is the outcome of iterative discriminative optimization against cross-entropy loss, which is free, in principle, to converge to any direction that separates the training folds well, including one that overfits idiosyncratic structure orthogonal to the class means. If the probe were exploiting such structure, we would expect \(\hat w_\ell\) to diverge from \(\hat d_\ell\) even while both achieved high held-out AUROC. The observed cosine of \(0.8495\) on average -- with alignment exceeding \(0.95\) in the earliest layers and remaining above \(0.75\) even at its minimum near layer 15 (Figure~\ref{fig:rq1-alignment}) -- indicates the discriminative optimum is close to the direction implied by simple class statistics. Agreement between two independently derived estimators is stronger evidence for a genuinely low-dimensional separating structure than either estimator's held-out performance alone.

\subsubsection{Layer-wise Behavior: Early, Middle, and Late Layers}

The layer-wise alignment and AUROC curves (Figures~\ref{fig:rq1-alignment} and \ref{fig:rq1-single-direction}) exhibit three qualitatively distinct regimes, described here as an empirical pattern without causal or mechanistic interpretation.

\emph{Early layers (\(\ell\lesssim 5\)).} Direction alignment is highest here (\(\approx 0.99\) at layer 1), while one-dimensional AUROC is already high (\(\approx 0.99\)) but not yet at ceiling. A plausible, non-causal reading is that early-layer representations retain substantial surface-form information -- which correlates with the reasoning/factual-recall split via vocabulary and format differences between source datasets -- and both estimators initially latch onto this shared surface signal.

\emph{Middle layers (\(\ell\approx 10\text{--}16\)).} Alignment reaches its minimum here (as low as \(\approx 0.73\) near layer 15), even though AUROC for both one-dimensional constructions is already at or near \(1.0\). This is the most informative regime: near-perfect classification does not imply a uniquely determined separating direction. Multiple distinct directions can each achieve ceiling AUROC, and the two estimators evidently select slightly different ones from this near-equivalent set.

\emph{Late layers (\(\ell\gtrsim 20\)).} Alignment partially recovers (toward \(0.85\)--\(0.89\)) while AUROC remains at ceiling throughout, consistent with the representation reorganizing toward a narrower, task-specific subspace as information is consolidated, though we treat this only as a descriptive observation.

None of these layer-wise observations license inferences about \emph{what computation} occurs at a given layer; they describe only the geometry of \(d_\ell\), \(\hat w_\ell\), and their alignment as a function of depth.

At the best mean-difference layer (layer 18), held-out projections are almost disjoint. Factual-recall examples have mean projected score approximately \(-19.73\), while reasoning-oriented examples have mean approximately \(19.73\), giving an absolute mean separation of \(39.46\). The reported overlap interval is only approximately \(0.0675\).

\rqfigure{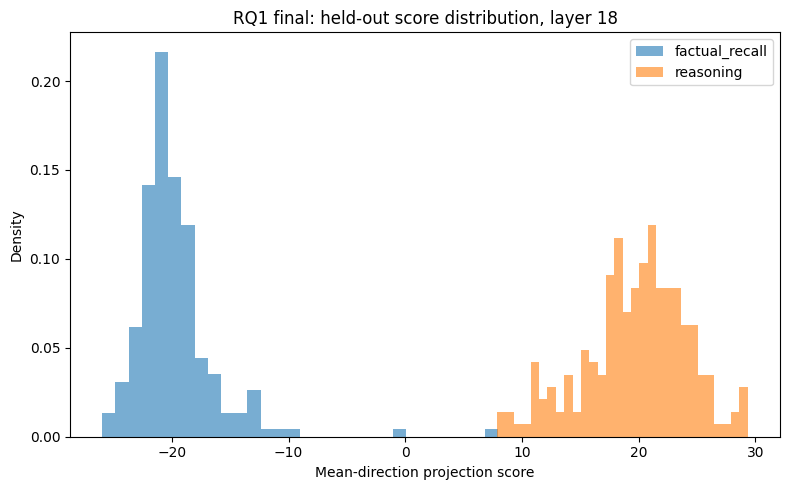}
{Held-out layer-18 projection scores along the mean-difference direction. The two task groups occupy almost disjoint scalar ranges.}
{fig:rq1-score}

\subsection{RQ1 Interpretation}

RQ1 therefore gives a qualified positive answer. For this model and controlled dataset, the reasoning-oriented/factual-recall distinction is not merely linearly separable: \textbf{one scalar direction is sufficient to match the full 1024-D probe}. This is consistent with the fixed-direction premise at a single model state.

But RQ1 does not establish that the direction represents an intrinsic reasoning mechanism. The strong TF--IDF baseline demonstrates that task-group labels remain heavily correlated with lexical/domain properties. The correct conclusion is therefore:

\begin{keybox}
\textbf{RQ1 conclusion:} the studied task groups admit an approximately single-direction representation in Qwen3-0.6B, but the direction should be interpreted as an empirical task-group separator rather than a causal ``reasoning neuron'' or universal semantic axis.
\end{keybox}

\section{RQ2: Representation Stability After GRPO}

RQ2 tests the stronger assumption required by a fixed-direction interpretation: that a direction extracted at one model state remains geometrically meaningful after RL changes the model parameters.

\subsection{GRPO Reorganizes the Separating Direction}

The base-to-GRPO direction cosines are substantially below one. The mean-direction cosine has
\[
\mu=0.4534,\quad
\sigma=0.1794,\quad
\text{median}=0.4919,
\]
with a minimum of \(0.1734\). The independently learned probe-direction cosine has
\[
\mu=0.4448,\quad
\sigma=0.1820,\quad
\text{median}=0.4393,
\]
with a minimum of \(0.1728\).

The layer profile is informative. Similarity is high in the earliest layers, falls sharply through the middle of the network, reaches its minimum around layer 15, and partially recovers in later layers. Thus the change is not a uniform scaling effect; it is a \textbf{layer-dependent geometric reorganization}.

\rqfigure{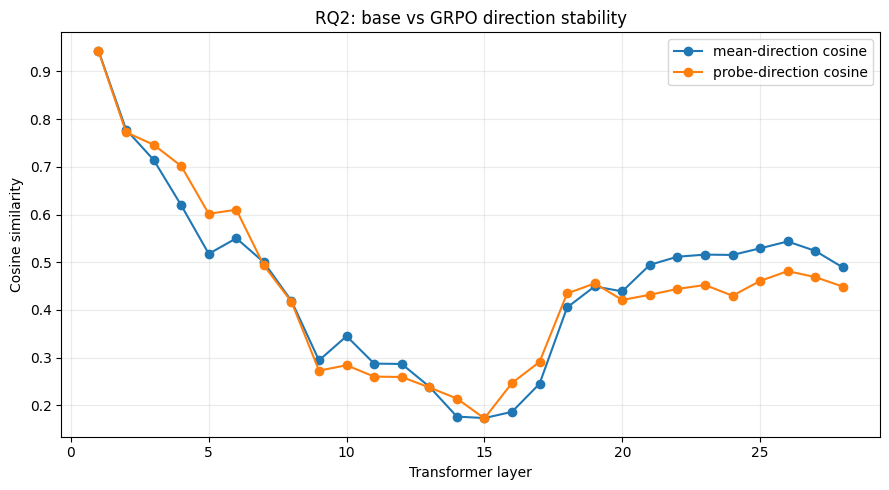}
{Base-to-GRPO cosine similarity for the mean-difference and independently learned probe directions. Both measures show strong early alignment followed by substantial rotation in middle layers and partial recovery in later layers.}
{fig:rq2-direction}

\subsection{Representation Drift Increases with Depth}

Direct comparison of corresponding hidden states gives a complementary result. Mean drift across layers is approximately \(0.2409\). At layer 1, cosine similarity is approximately \(0.9872\), corresponding to only \(0.0128\) drift. Drift increases substantially through the network and reaches its largest value at layer 28, where cosine similarity is approximately \(0.4894\), corresponding to \(0.5106\) drift.

\rqfigure{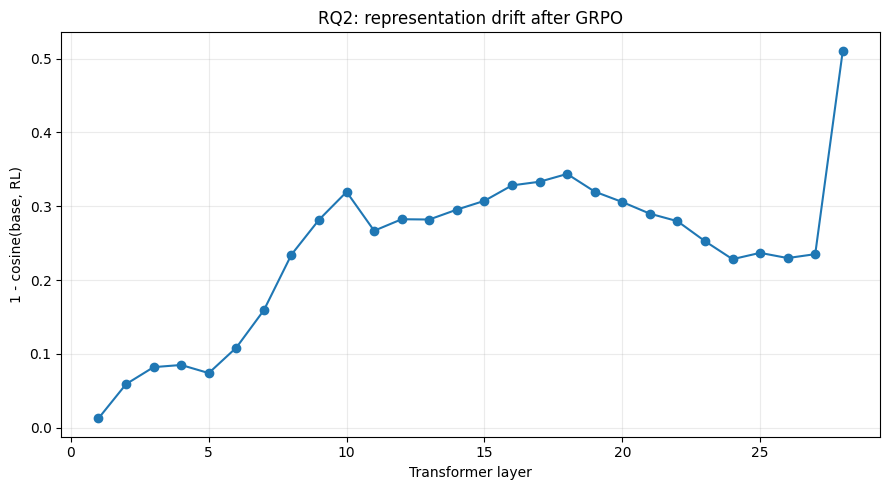}
{Representation drift between the base and GRPO models. Early layers remain highly aligned, whereas deeper layers undergo substantially larger geometric changes.}
{fig:rq2-drift}

The important point is that \textbf{directional reorganization and broader representation drift agree}. The effect is therefore not confined to a single classifier vector.

\subsection{Decodability Survives the Reorganization}

Despite this geometric change, the task-group distinction remains almost perfectly decodable after GRPO. The best base-model layer reaches \(0.9975\) accuracy and \(1.0000\) AUROC; the corresponding RL checkpoint reaches \(0.9950\) accuracy and \(1.0000\) AUROC.

\rqfigure{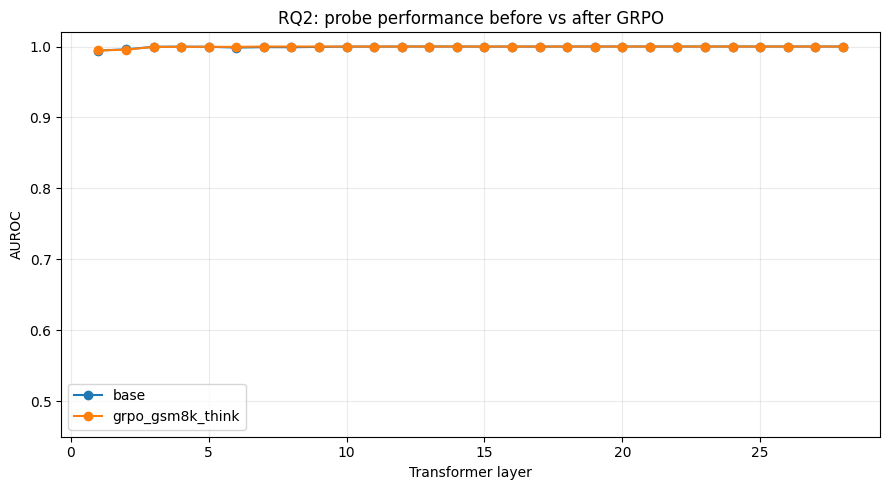}
{Layer-wise probe AUROC before and after GRPO. Near-perfect decoding survives despite substantial changes in the separating geometry.}
{fig:rq2-probe}

This is the key contrast of the paper:

\[
\boxed{
\text{Decodability}\approx\text{preserved}
\qquad\neq\qquad
\text{Direction}\approx\text{preserved}.
}
\]

The result is consistent with a \textbf{representation reparameterization} interpretation. GRPO does not erase the information required to distinguish the two task groups, but it changes the geometric coordinates in which that information is expressed.

\subsection{Information Preservation Does Not Imply Geometric Preservation}
\label{sec:info-vs-geometry}

This subsection is the conceptual centerpiece of RQ2, and it is worth stating the apparent tension explicitly before resolving it.

\begin{quote}
\emph{How can the base-to-RL direction cosine average only \(0.453\), while a probe trained on the RL representation still reaches AUROC \(=1.0\)?}
\end{quote}

If the reasoning/factual-recall direction were a single, mechanistically fixed axis, cosine similarity near \(0.45\) -- roughly midway between orthogonality (\(0\)) and perfect alignment (\(1\)) -- would be difficult to reconcile with essentially unchanged decodability. The resolution is that \emph{decodability} and \emph{direction stability} are measuring different mathematical objects, and only one of them is invariant to rotation of the representation.

Held-out AUROC evaluates whether \emph{some} linear direction separates the two classes in the RL representation; it says nothing about whether that direction coincides with the one that separated them in the base representation. Formally, let \(\hat d_\ell^{B}\) and \(\hat d_\ell^{RL}\) be the respective mean-difference directions. AUROC computed from \(\hat d_\ell^{RL}\) on RL-model representations depends only on the RL representation and the RL-fitted direction; it is entirely insensitive to \(\operatorname{cos}(\hat d_\ell^{B},\hat d_\ell^{RL})\). It is therefore mathematically possible -- and, empirically, exactly what we observe -- for this cosine to be far below \(1\) while both \(\hat d_\ell^{B}\) and \(\hat d_\ell^{RL}\), each evaluated on their own checkpoint, independently achieve ceiling AUROC.

A useful analogy is a rigid object viewed under two different camera angles: the object's identity (linear separability of the two task groups) is preserved, but the coordinate axes describing its orientation (the specific direction vector) are not. GRPO applies a large, layer-dependent rotation to the representation manifold without necessarily destroying the \emph{separation} between task groups -- and our results indicate that, for this checkpoint, it does not.

We therefore treat \(0.453\) mean-direction cosine and \(1.0\) post-RL AUROC as two measurements of orthogonal properties, not conflicting evidence: one of a fixed vector's persistence, the other of the existence of \emph{some} separating vector. Conflating them -- treating high post-RL decodability as evidence that the \emph{original} direction remains valid -- is exactly the inferential error a fixed-direction RL method could make if its anchor were assumed, rather than measured, to be stable.

\subsubsection{Why Drift and Rotation Are Complementary, Not Redundant}

Direction stability (Section~\ref{sec:formulation}) and representation drift (Section~\ref{sec:repr-drift}) could, in principle, disagree. Direction stability is computed only from the task-relevant direction estimators \(\hat d_\ell\) and \(\hat w_\ell\), which are themselves derived from class-averaged or discriminatively optimized statistics; a model could in principle preserve this specific task-relevant axis while broadly reorganizing every other aspect of its representation, or conversely could preserve most of the representation's geometry while specifically rotating the task-relevant axis. Representation drift, by contrast, is computed per example from raw hidden states without reference to the task labels at all, and therefore captures reorganization that is invisible to a label-dependent direction estimator.

The fact that both quantities show the same qualitative layer-wise pattern -- high stability/low drift in early layers, a pronounced dip/spike through the middle and late layers -- is itself an empirical finding, not a logical necessity. It indicates that the geometric reorganization induced by GRPO is not confined to the specific reasoning/factual-recall axis; it is a broader property of how GRPO reshapes the representation at each depth. This strengthens the RQ2 conclusion: the observed direction rotation is not an artifact of how \(d_\ell\) or \(\hat w_\ell\) happen to be estimated, since an unrelated, label-free measurement of representational change corroborates it.

\subsection{Independent Probe-Direction Check}

The same conclusion is obtained when the representation is characterized by independently trained probe vectors rather than class-mean differences. The probe-direction cosine follows the same broad layer-wise pattern as the mean-direction cosine, providing an important robustness check against defining the direction solely from class means.

\rqfigure{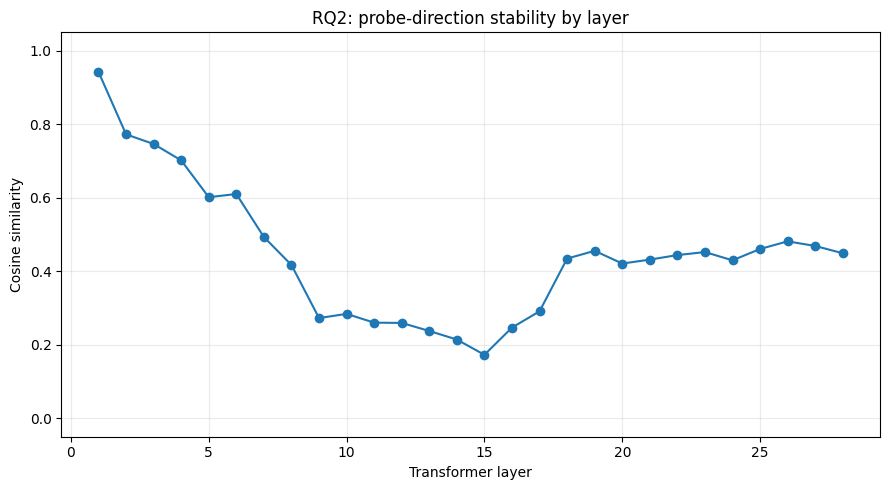}
{Cosine similarity between independently learned probe directions at corresponding base and GRPO layers. The classifier remains predictive even where the learned separating vector rotates.}
{fig:rq2-probe-direction}

\subsection{RQ2 Interpretation}

RQ2 does not support the stronger assumption that a direction extracted before RL is a training-invariant representation. Instead, it provides evidence for a weaker and more precise statement:

\begin{keybox}
\textbf{RQ2 conclusion:} GRPO preserves the decodability of the studied reasoning--factual-recall distinction while substantially reorganizing its representation geometry. A fixed direction is therefore an empirical anchor, not a guaranteed invariant of post-training.
\end{keybox}

\section{Joint Interpretation}

The two research questions produce a useful progression.

\paragraph{Phase I -- representation exists.}
RQ1 establishes that, for the controlled task groups, a single scalar projection can match a full 1024-dimensional probe. The evidence is therefore compatible with a low-dimensional representation of the task distinction.

\paragraph{Phase II -- representation moves.}
RQ2 then shows that this compact representation is not geometrically stationary under GRPO. Direction cosine drops to roughly \(0.45\) on average, while late-layer representation drift exceeds \(0.50\).

\paragraph{The combined result.}
The central scientific observation is not that ``reasoning is not a direction.'' In fact, RQ1 shows that a direction is highly effective for the studied task groups. Rather, the stronger claim that \emph{the same direction remains fixed throughout post-training} is not supported by the observed cross-model geometry.

This distinction matters for direction-aware RL methods such as DiRL. Our results do not show that a fixed direction cannot improve RL, nor do they evaluate DiRL's downstream reward or accuracy. They show that the representational stability assumed when interpreting a pre-RL direction as a persistent mechanistic object should be \textbf{measured rather than presumed}.

\subsection{Summary of Hypotheses and Findings}

Table~\ref{tab:hypotheses} consolidates the five empirical hypotheses tested across RQ1 and RQ2, together with the specific evidence bearing on each. Framing the results as discrete, individually falsifiable hypotheses makes explicit which claims are supported, which are contradicted, and which are simply outside the scope of the present experiments.

\begin{table}[t]
\centering
\small
\begin{tabular}{p{0.1cm}p{5.6cm}c}
\toprule
& Hypothesis & Verdict\\
\midrule
H1 & Reasoning-oriented and factual-recall examples are linearly separable in hidden-state space. & Supported\\
H2 & A single one-dimensional direction is sufficient to match a full 1024-D probe. & Supported\\
H3 & The mean-difference / probe direction extracted before RL remains the same direction after GRPO. & Not supported\\
H4 & The broader representation geometry (not only the task direction) is stationary under GRPO. & Not supported\\
H5 & Task-relevant information is preserved after GRPO even though its geometric encoding changes. & Supported\\
\bottomrule
\end{tabular}
\caption{Hypothesis-level summary. H1--H2 concern RQ1 (separability and dimensionality); H3--H5 concern RQ2 (stability). The asymmetric pattern -- H1, H2, and H5 supported, H3 and H4 not supported -- is the central empirical result of the paper.}
\label{tab:hypotheses}
\end{table}

The pattern in Table~\ref{tab:hypotheses} is deliberately asymmetric. H1 and H2 concern properties of a \emph{single} model state and are both supported by the RQ1 evidence in Section~\ref{sec:formulation} and Table~\ref{tab:rq1-main}. H3 and H4 concern properties that require \emph{comparing two model states} and are both contradicted by the RQ2 evidence in Section~\ref{sec:info-vs-geometry}. H5 again concerns a property that can be evaluated at each model state independently (decodability), and is supported at both states despite H3 and H4 failing. No single hypothesis in this table is self-contradictory with any other; the appearance of tension arises only if H2 (single-direction sufficiency \emph{at a state}) is mistaken for H3 (single-direction \emph{persistence across states}), which is exactly the conflation Section~\ref{sec:info-vs-geometry} addresses.

\section{Discussion}

\subsection{What the Results Support}

The experiments support three bounded conclusions.

First, the studied reasoning-oriented and factual-recall task groups are strongly distinguishable in Qwen3-0.6B hidden-state space. Second, this distinction is approximately one-dimensional under the final RQ1 protocol: the mean-difference and probe-weight directions reach the same AUROC as the full linear probe. Third, the distinction remains decodable after GRPO.

\subsection{What the Results Challenge}

The experiments challenge a stronger interpretation of a fixed direction as a stable internal object. If the direction were preserved, one would expect high cross-model cosine similarity at corresponding layers. Instead, both mean-difference and probe-direction comparisons show substantial rotation, particularly in middle layers.

The simultaneous presence of high AUROC and low-to-moderate directional cosine is not contradictory. A discriminative property can survive a change of coordinates. In other words, the model may preserve the \emph{information} while changing the \emph{basis} used to encode it.

\subsection{Why This Is a Representation Result, Not Yet a Mechanistic Causal Result}

We intentionally use ``representation analysis'' rather than claiming causal mechanistic discovery. Cosine similarity measures geometry, not computation. A rotated representation can still implement the same underlying algorithm, and a stable direction can in principle be causally irrelevant.

A causal claim would require interventions such as activation patching, targeted representation ablation, causal mediation, or related techniques. Those experiments are outside the scope of this paper.

\subsection{Implication for Adaptive Representation Methods}

The result suggests a practical design principle for future representation-aware post-training: if a representation is used as an optimization signal, it may be preferable to \textbf{re-estimate or track the relevant subspace during training} rather than treating a single pre-training direction as permanently valid.

This is a motivation, not an evaluated algorithm. We deliberately leave adaptive tracking as future work.

\subsection{What This Means, and Does Not Mean, for DiRL}
\label{sec:dirl-discussion}

Because DiRL~\citep{xia2026dirl} is the most direct downstream consumer of the fixed-direction assumption tested here, we discuss its relationship to our results carefully and separately from the general discussion above.

It is important to distinguish two claims that are easy to conflate:

\begin{enumerate}[leftmargin=1.35em,itemsep=1pt,topsep=2pt]
    \item[(a)] \key{Operational usefulness:} a direction extracted before RL can be a useful exploration signal or reward-shaping anchor during training, regardless of whether it remains geometrically aligned with the representation at the end of training.
    \item[(b)] \key{Mechanistic invariance:} the same direction continues to correspond to the reasoning/factual-recall distinction, in the same coordinate frame, throughout and after training.
\end{enumerate}

Our experiment speaks only to claim (b), and only for the specific base-to-GRPO transition studied here. We do not run DiRL, measure its downstream reward, or compare models trained with and without a fixed-direction exploration bonus. Consequently, \textbf{nothing in our results implies that DiRL's fixed-anchor design is empirically inferior, or that DiRL fails to improve reasoning performance}. A fixed anchor can remain operationally useful throughout training precisely because it does not need to stay a mechanistically privileged axis to bias the policy usefully early on.

What our results do imply is narrower: interpreting a DiRL-style anchor as a \emph{persistent mechanistic object} -- e.g.\ using the post-training projection onto the original anchor as evidence the model is ``still reasoning along the same axis'' -- is not supported without directly measuring post-training direction stability, since stability is not automatic even when decodability is perfectly preserved. We recommend that future direction-aware RL work report post-training direction cosine and representation drift alongside downstream performance, so that (a) and (b) are assessed as the separate empirical questions they are.

\section{Limitations}

\begin{itemize}[leftmargin=1.25em,itemsep=2pt,topsep=2pt]
    \item \key{One model family.} All experiments use Qwen3-0.6B. Whether the findings generalize to larger Qwen3 variants or other model families is untested; direction rotation speed under RL could plausibly depend on scale or architecture.
    \item \key{One RL state.} RQ2 compares the base model with one publicly released GRPO checkpoint rather than a controlled trajectory, so we cannot report \emph{when} during training rotation occurs or whether it would plateau or reverse with further training.
    \item \key{External RL checkpoint.} The RL checkpoint was not trained under our protocol, so we cannot attribute the observed geometry to a specific hyperparameter or training trajectory (reward design, KL penalty, learning rate, step count).
    \item \key{Task-group confounding.} TF--IDF remains highly predictive (Table~\ref{tab:baselines}), so lexical and domain structure are not eliminated by length matching. Matching additionally on TF--IDF score could reduce this confound further, at the cost of additional data loss.
    \item \key{Operational label.} Factual recall is used as a practical comparison class, not assumed identical to memorization; some questions may be answerable through inference rather than pure retrieval, and some reasoning problems may be solvable via memorized templates.
    \item \key{Extraction position.} The main analysis uses the final formatted prompt-token representation; intermediate token positions, generated-token representations during chain-of-thought, and attention-pattern signals are not examined.
    \item \key{Cosine similarity as a stability metric.} Cosine similarity does not distinguish a rotation specific to the task-relevant subspace from a global change of basis affecting all directions equally; disentangling these would require a null distribution over task-irrelevant directions, which we do not construct.
    \item \key{Linear probing assumption.} Both direction estimators assume an approximately linear encoding; a non-linear probe could reveal additional structure, though near-perfect linear AUROC leaves little practical headroom to improve on.
    \item \key{No causal claim.} Representation drift and directional rotation do not establish a changed reasoning mechanism -- a rotated direction is a geometric description, not a demonstration of changed computation.
    \item \key{No downstream RL evaluation.} We do not test whether recomputing the direction would improve GRPO, or whether a fixed direction is empirically inferior for RL (Section~\ref{sec:dirl-discussion}).
\end{itemize}

These limitations define the scope of the claims and are important when comparing the result with broader claims about reasoning representations.

\section{Reproducibility}

The complete analysis is designed around deterministic, inspectable artifacts. The principal representation tensors have shape \([400,28,1024]\). Saved outputs include base and RL hidden states, final-layer representations, layer-wise direction cosines, representation-drift measurements, probe results, and generated figures.

The RQ2 comparison uses the public checkpoint
\texttt{x32/Qwen3-0.6B-GRPO-GSM8K-Think}. No claim of having trained this checkpoint is made.

\subsection{Configuration Details}

\paragraph{Model and tokenizer.} Both comparisons use the Qwen3-0.6B architecture and its associated tokenizer \citep{qwen3}; the RL checkpoint is loaded from the publicly released \texttt{x32/Qwen3-0.6B-GRPO-GSM8K-Think} weights without modification, and hidden states are extracted in evaluation mode (no dropout) for deterministic representations.

\paragraph{Prompt template.} All 400 examples are passed through a single common formatting template before tokenization, so that differences in raw dataset formatting do not introduce a spurious source of separability. Token lengths reported throughout the paper are measured \emph{after} this formatting step.

\paragraph{Representation tensor and CV protocol.} The principal representation tensors have shape \([400,28,1024]\), saved separately for the base and RL checkpoints with examples indexed in the same fixed order, which is what makes the per-example drift computation (Section~\ref{sec:repr-drift}) well defined. All held-out accuracy and AUROC figures are computed with 5-fold stratified cross-validation on the reasoning/factual-recall label, preserving the 200/200 class balance in each fold.

The main reproducibility checklist is:

\begin{itemize}[leftmargin=1.25em,itemsep=1pt,topsep=2pt]
    \item model and tokenizer configuration;
    \item common prompt template;
    \item dataset source and sample metadata;
    \item token-length matching procedure;
    \item exact representation extraction position;
    \item layer-wise hidden-state extraction;
    \item 5-fold stratified probing;
    \item mean-difference and probe-direction construction;
    \item cosine-based stability analysis;
    \item representation-drift computation;
    \item figure-generation scripts and saved numerical outputs.
\end{itemize}

\subsection{Saved Artifacts and Figure Paths}

Intermediate outputs are saved at each pipeline stage rather than only the final figures: raw and deduplicated dataset metadata; the length-binned 400-example index; base and RL hidden-state tensors; per-layer direction vectors for both checkpoints; per-layer cosine-similarity and drift scalars; and the cross-validated accuracy/AUROC numbers underlying every table. Figures are generated from these saved numerical outputs rather than recomputed inline, so a given figure can be regenerated without rerunning representation extraction or probing. This separation between raw representations, numerical results, and rendered figures makes the analysis straightforward to reproduce without rerunning unrelated experiments.

\section{Conclusion}

We investigated whether reasoning-oriented and factual-recall representations in Qwen3-0.6B can be characterized by a single direction and whether that direction remains stable after reinforcement learning.

The answer is asymmetric. \textbf{RQ1 supports single-direction decodability}: the best mean-difference direction and probe-weight direction both reach AUROC \(=1.00\), matching the full 1024-dimensional probe on the controlled task groups. \textbf{RQ2 challenges fixed-direction stability}: after GRPO, mean-direction cosine averages only \(0.453\), probe-direction cosine \(0.445\), and final-layer representation drift reaches approximately \(0.511\), while probe AUROC remains \(1.00\).

The resulting picture is therefore not that reasoning cannot be represented by a direction. Rather, the evidence suggests that \textbf{a direction can be a compact representation at one model state without being invariant under post-training}. This distinction is important for interpreting direction-aware RL and representation-based mechanistic analyses.

We conclude with a deliberately modest claim: \textbf{representation-level information can remain stable while representation geometry changes}. Treating a pre-RL direction as fixed should therefore be regarded as an empirical assumption to validate, not as a mechanistic invariant.

\appendix

\section{Original MATH-500 versus PopQA Diagnostic}
\label{app:original-diagnostic}

This appendix documents the two earlier stages of dataset construction described in Section~3.2, both of which motivated -- but were explicitly not used as evidence for -- the final RQ1/RQ2 protocol reported in the main text.

\subsection{Stage 1: The Unmatched Diagnostic}

The original diagnostic compared MATH-500 \citep{hendrycks2021math} directly against PopQA \citep{mallen2023popqa}, labeled here as ``Reasoning'' and ``Memorization'' following the terminology of the initial exploratory analysis before we adopted the more careful \emph{factual recall} label used in the main text. Figure~\ref{fig:app-original-pca} shows the final-layer PCA for this unmatched comparison: the two groups separate cleanly along the first principal component, but this is confounded with the pronounced token-length mismatch shown in Figure~\ref{fig:app-strict-before}. Under this construction, a length-only classifier reaches approximately \(0.965\) accuracy and \(0.994\) AUROC -- too high for hidden-state separability to be attributed to task-semantic content alone.

\rqfigure{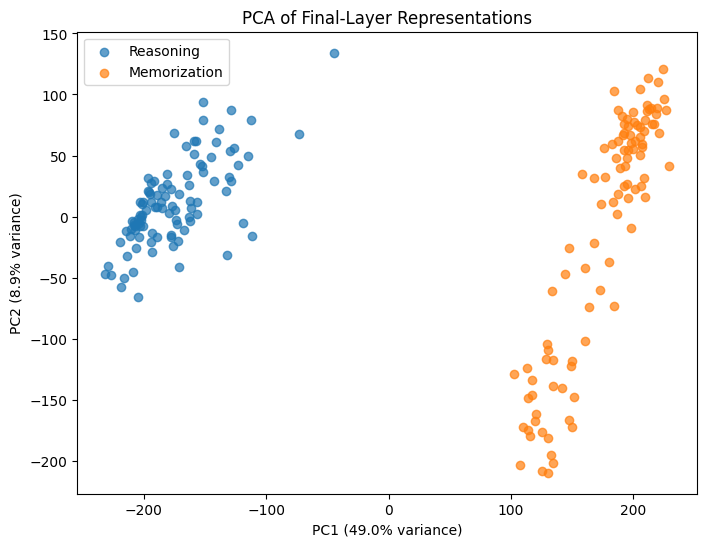}
{Final-layer PCA of the original, unmatched MATH-500-versus-PopQA diagnostic. The clean linear separation visible here is consistent with -- but does not on its own distinguish between -- a semantic reasoning/memorization signal and a token-length confound.}
{fig:app-original-pca}

\rqfigure{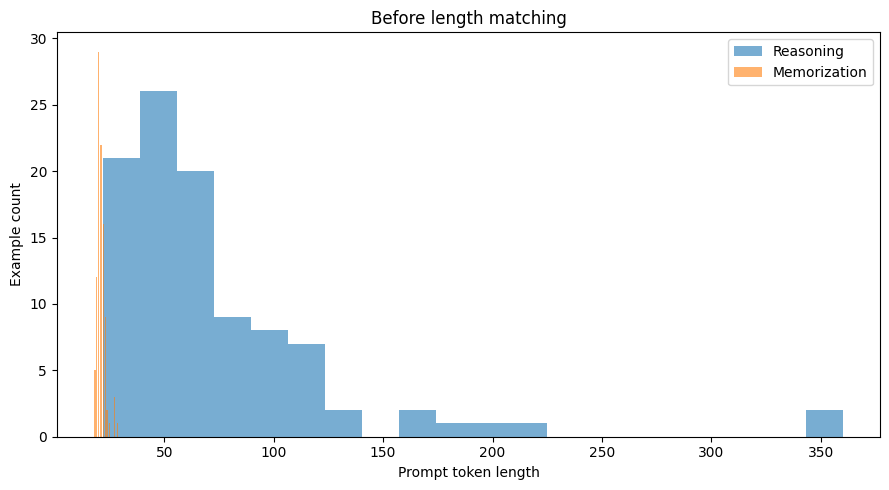}
{Prompt token-length distributions for the original MATH-500-versus-PopQA diagnostic before any length control. The Memorization (PopQA) group is concentrated at very short token lengths while the Reasoning (MATH-500) group spans a much wider and longer range, motivating the length-matching procedures described in Section~3.2.}
{fig:app-strict-before}

\subsection{Stage 2: Strict Pairwise Length Matching}
\label{app:length-diagnostics}

To directly test whether the Stage 1 separation survives length control, we constructed a strict pairwise length-matched subset in which each reasoning example is matched to a factual-recall example of nearly identical token length. Figure~\ref{fig:app-strict-pca} shows the resulting before/after token-length histograms: the matched subset achieves near-exact length equalization, but at a steep cost in sample size, yielding only 20 usable examples. Figure~\ref{fig:app-strict-final-pca} shows the corresponding final-layer PCA; the two groups remain visually separable, a useful qualitative signal but not a statistically adequate basis for the held-out cross-validated claims in the main text -- precisely why we adopted the expanded, binned-matching construction (Section~3.2, Stage 3) instead.

\rqfigure{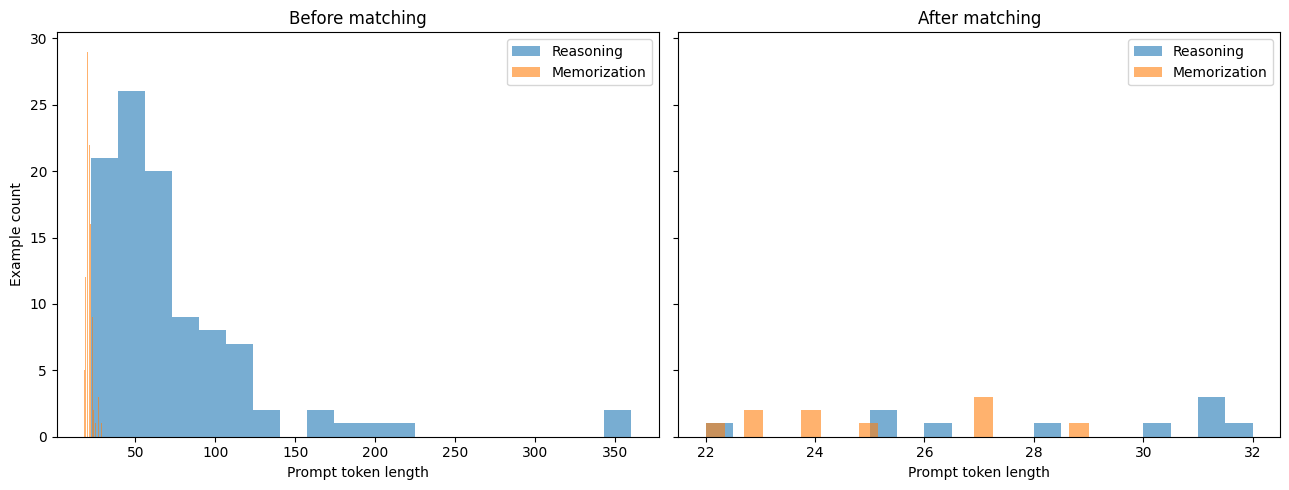}
{Prompt token-length distributions before (left) and after (right) strict pairwise length matching in the original diagnostic. The right panel shows near-exact length equalization between the Reasoning and Memorization groups, but only 20 examples remain after this strict matching procedure.}
{fig:app-strict-pca}

\rqfigure{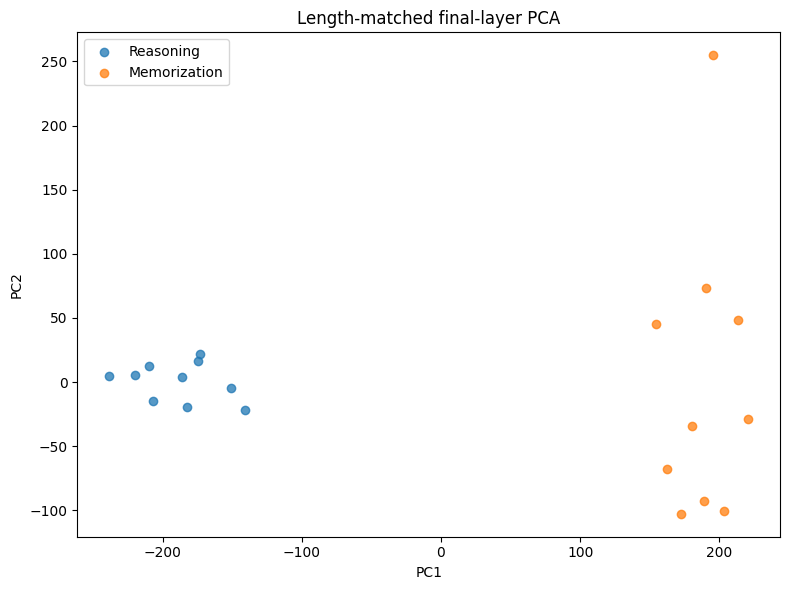}
{Final-layer PCA of the strict, pairwise length-matched diagnostic (\(n=20\)). The two groups remain qualitatively separable even under strict length control, providing early motivation for the expanded, better-powered construction used in the main text.}
{fig:app-strict-final-pca}

\section{Additional RQ1 Diagnostics}

The following figure is retained as a supporting diagnostic rather than central evidence for the final, expanded-dataset RQ1 protocol.

\rqfigure{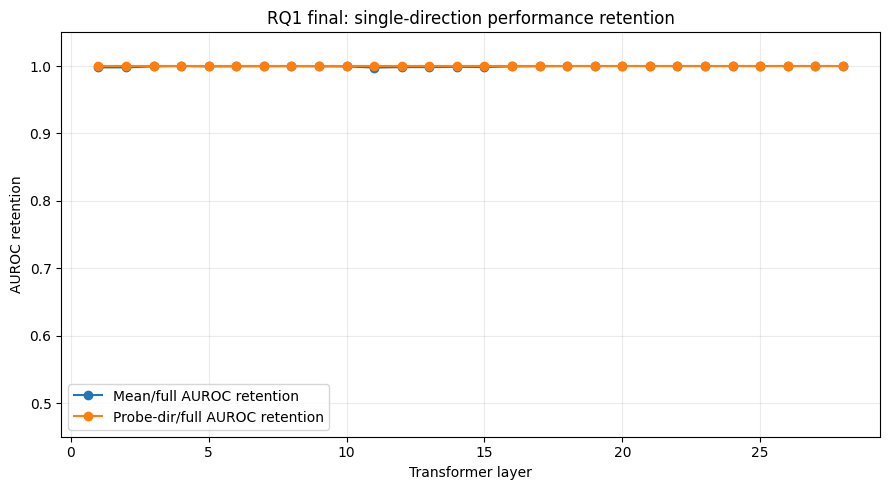}
{Retention of one-dimensional AUROC relative to the full probe. The one-dimensional directions retain essentially all probe performance.}
{fig:app-retention}

\paragraph{Original versus expanded controls.}
The initial dataset comparison was deliberately not used as the final scientific evidence because token length alone predicted the label very strongly. The expanded control reduces this confound but does not eliminate lexical/domain confounding.

\rqfigure{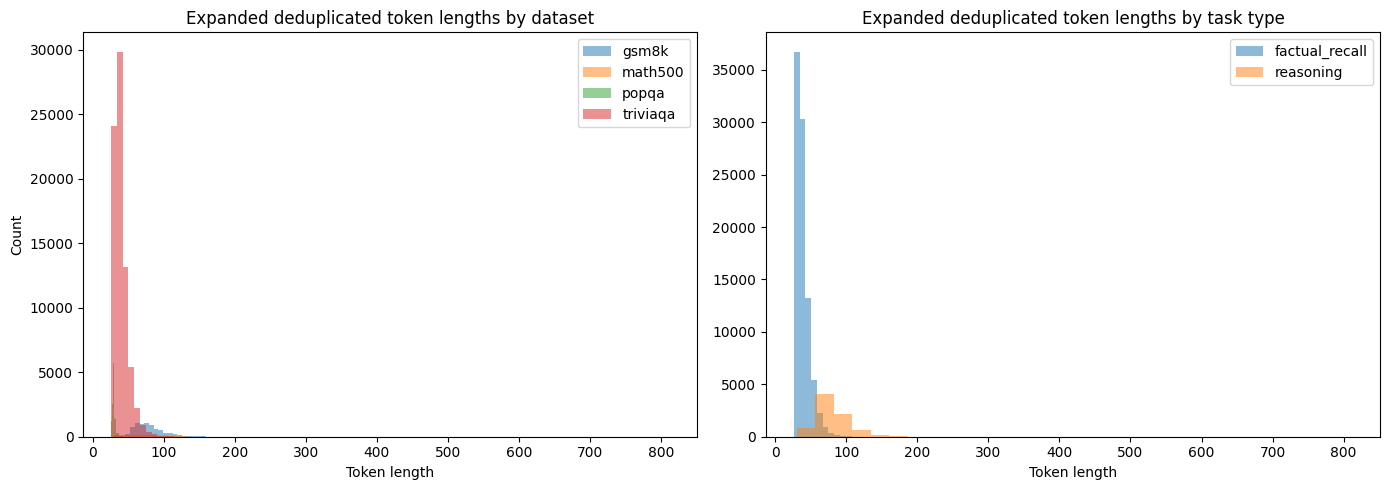}
{Expanded deduplicated token-length distributions by source dataset (left) and by task type (right), prior to length-bin matching. This diagnostic motivates explicit length control before interpreting hidden-state separation.}
{fig:app-length-datasets}

\rqfigure{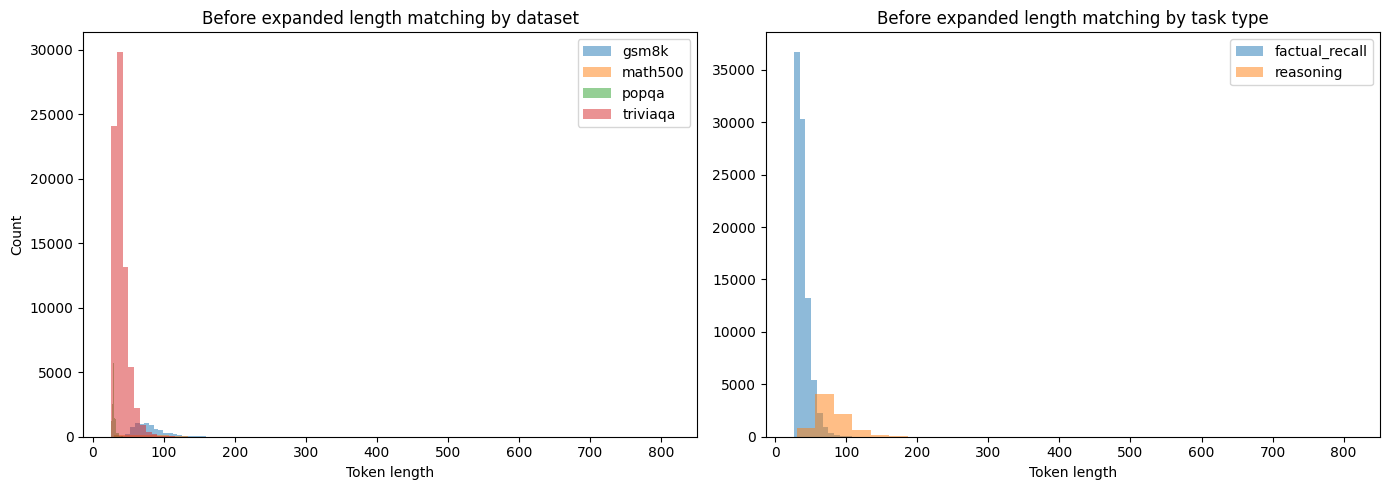}
{Token-length distributions by source dataset (left) and by task type (right) before expanded length matching. The strong mismatch illustrates why the uncorrected dataset cannot support a clean representation claim.}
{fig:app-length-before}

\section{Probe and Checkpoint Protocol Details}

\subsection{Logistic-Regression Probe Protocol}

At each of the 28 layers, we fit a separate logistic-regression probe \(w_\ell\in\mathbb{R}^{1024}\) on the training folds of each cross-validation split. Probes are fit independently per layer and per checkpoint (base or RL) -- no parameter sharing across layers or across the base/RL comparison -- so that any observed similarity between base-model and RL-model probe directions (Section~5.4) reflects a genuine property of the two representations. The full 1024-dimensional probe (Table~\ref{tab:rq1-main}, Figure~\ref{fig:rq1-full-probe}) uses the same procedure and cross-validation splits as the one-dimensional constructions, so all classifiers in Table~\ref{tab:baselines} are evaluated under identical held-out conditions.

\subsection{Random-Label and Lexical Control Protocol}

The random-label probe (Table~\ref{tab:baselines}) is trained identically to the full probe but with labels randomly shuffled prior to fitting; its near-chance accuracy (\(0.4875\)) and AUROC (\(0.4816\)) confirm the pipeline does not manufacture separability from noise alone. The TF--IDF baseline is fit on bag-of-words features over the formatted prompt text with the same cross-validation protocol, so its numbers in Table~\ref{tab:baselines} are directly comparable to the hidden-state-derived ones.

\subsection{RL Checkpoint Details}

The RQ2 comparison uses the publicly released checkpoint \texttt{x32/Qwen3-0.6B-GRPO-GSM8K-Think}, which applies GRPO post-training with a GSM8K-derived reward to the Qwen3-0.6B base model. We did not train this checkpoint and make no claim about its specific hyperparameters or training trajectory; it is treated strictly as an externally available post-training state. Both checkpoints are evaluated on the identical 400-example dataset and prompt template, so any observed cosine or drift difference reflects a genuine representational change rather than a difference in input distribution.

\bibliographystyle{plainnat}
\bibliography{references}

\end{document}